\documentclass[Crown,sagev,times]{sagej}

\usepackage{moreverb,url}
\usepackage[colorlinks,bookmarksopen,bookmarksnumbered,citecolor=red,urlcolor=red]{hyperref}
\usepackage{soul}\setuldepth{article}
\usepackage{tabularx}
\usepackage{array}
\usepackage{booktabs}
\usepackage{xcolor}
\usepackage{listings}
\usepackage{caption}
\usepackage{graphicx}
\usepackage{algorithm}
\usepackage{algpseudocode}
\usepackage{amsmath}
\usepackage{multirow}
\usepackage{tikz}
\usetikzlibrary{shapes.geometric, arrows}
\usepackage[table]{xcolor}
\definecolor{TNOBlue}{HTML}{a0b2e2}
\definecolor{TNOOrange}{HTML}{fbc6a3}

\newcommand\BibTeX{{\rmfamily B\kern-.05em \textsc{i\kern-.025em b}\kern-.08em
T\kern-.1667em\lower.7ex\hbox{E}\kern-.125emX}}

\def\volumeyear{2016}

\usepackage{mathptmx}

\newcommand{\triple}[1]{{\small\textsc{#1}}}

\def\hb{\hbox to 11.5 cm{}}

\begin{document}

\runninghead{Bakker et al.}

\title{Ontology Learning with LLMs: A Benchmark Study on Axiom Identification}



\author{Roos M. Bakker\affilnum{1,2}, Daan L. Di Scala\affilnum{1,3}, Maaike H.T. de Boer\affilnum{1}, Stephan A. Raaijmakers\affilnum{1,2}}

\affiliation{\affilnum{1}Netherlands Organisation for Applied Scientific Research (TNO)
\\
\affilnum{2}Leiden University Centre for Linguistics (LUCL)\\
\affilnum{3}Utrecht University}

\corrauth{Roos M. Bakker, TNO
Anna van Buerenplein 1,
Den Haag,
2595 DA, NL.}

\email{roos.bakker@tno.nl}

\begin{abstract}
Ontologies are an important tool for structuring domain knowledge, but their development is a complex task that requires significant modelling and domain expertise. Ontology learning, aimed at automating this process, has seen advancements in the past decade with the improvement of Natural Language Processing techniques, and especially with the recent growth of Large Language Models (LLMs). This paper investigates the challenge of identifying axioms: fundamental ontology components that define logical relations between classes and properties. In this work, we introduce an Ontology Axiom Benchmark \textit{OntoAxiom}, and systematically test LLMs on that benchmark for axiom identification, evaluating different prompting strategies, ontologies, and axiom types. The benchmark consists of nine medium-sized ontologies with together 17.118 triples, and 2.771 axioms. We focus on subclass, disjoint, subproperty, domain, and range axioms. To evaluate LLM performance, we compare twelve LLMs with three shot settings and two prompting strategies: a Direct approach where we query all axioms at once, versus an Axiom-by-Axiom (AbA) approach, where each prompt queries for one axiom only. Our findings show that the AbA prompting leads to higher F1 scores than the direct approach. However, performance varies across axioms, suggesting that certain axioms are more challenging to identify. The domain also influences performance: the FOAF ontology achieves a score of 0.642 for the subclass axiom, while the music ontology reaches only 0.218. Larger LLMs outperform smaller ones, but smaller models may still be viable for resource-constrained settings. Although performance overall is not high enough to fully automate axiom identification, LLMs can provide valuable candidate axioms to support ontology engineers with the development and refinement of ontologies.
\end{abstract}


\maketitle

\section{Introduction}


Knowledge graphs and formal ontologies are increasingly vital in modern information systems and AI-based decision support systems. These formal representations provide a transparent and explainable complement to black-box machine learning models. Ontologies define type entities and relations that describe and structure aspects of a domain of interest \cite{studer1998knowledge}, enabling reasoning and interoperability over various data sources. However, constructing and maintaining such formal models is a complex and resource-intensive task, requiring manual effort not only for the initial development but also for updates to ensure alignment with evolving applications. Data-driven techniques can help automate this process by identifying ontology elements from text or other domain sources. In this paper, we investigate the ability of language models to automatically identify axioms in ontologies. 

\begin{figure}[ht]
    \centering
    \includegraphics[width=0.9\linewidth]{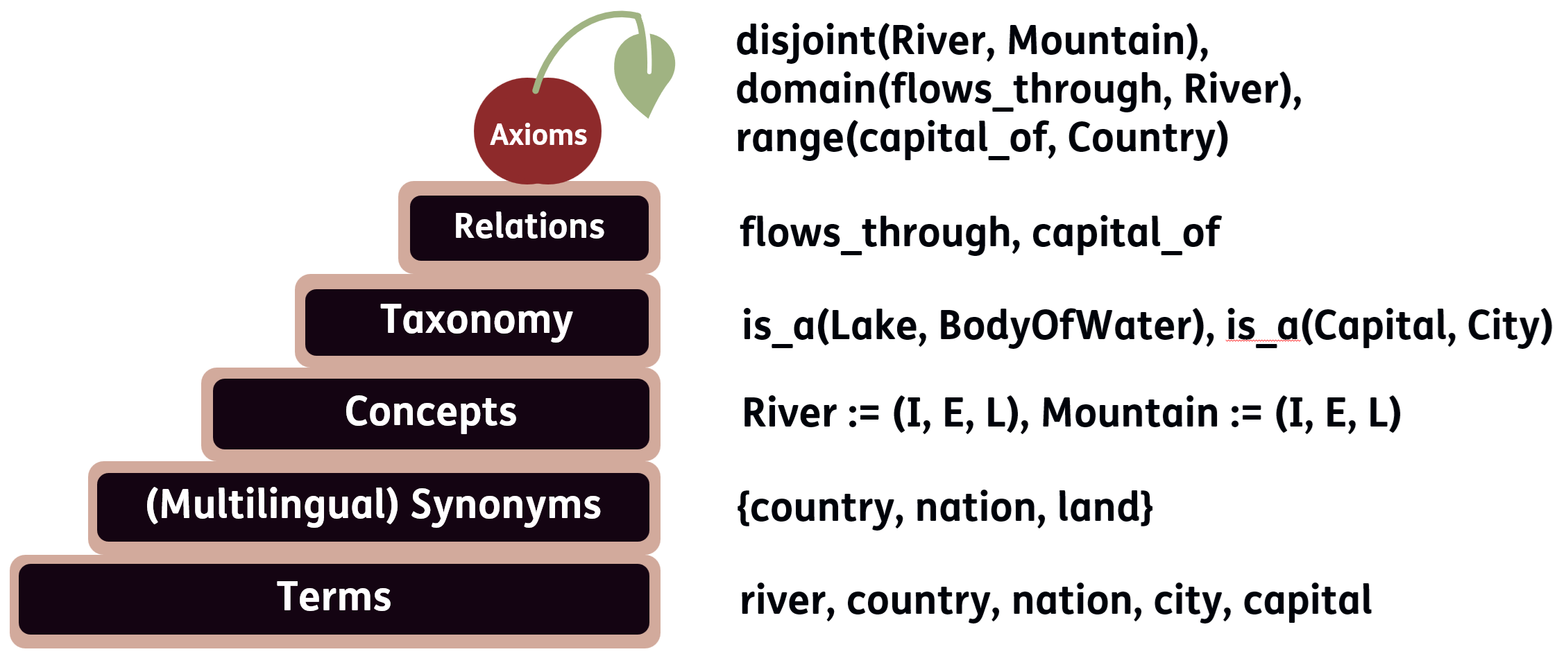}
    \caption{Ontology learning layer cake (Figure adapted from \cite{buitelaar2005ontologylearning}).}
    \label{fig:layer-cake}
\end{figure}

Early work on automatically identifying ontologies, also called ontology learning, made the distinction between different components of the ontology, such as terms, taxonomical relations, and axioms \cite{maedche2004ontologylearning,shamsfard2003state}. An example of an axiom is the disjoint relation, where two concepts are disjoint if no individual can simultaneously belong to both classes (e.g., a mountain cannot also be a river, and vice versa).  An early overview of Buitelaar et al. \cite{buitelaar2005ontologylearning} on ontology learning from text identified differences in complexity between these tasks and visualises it in the ontology learning layer cake, as shown in Figure \ref{fig:layer-cake}. At the bottom of the cake is the automatic learning of terms, considered the least complex task. It builds up to synonyms, concepts (consisting of intensions $I$, extensions $E$ and lexicon $L$) and taxonomical concept hierarchies. At the top layer are rules and axioms, with examples such as disjoint, domain, and range relations. A domain axiom specifies the allowed subject of a relation. For instance, the relation \triple{flows\_through} has a domain of \triple{river}. A range axiom specifies the allowed object of a relation, such as \triple{capital\_of}, which has a range of \triple{country}. Early research mainly focused on the lower levels by creating rule-based solutions using syntactical patterns, with the results needing manual adjustments to produce ontologies. In the last decade, the field of ontology learning and engineering has advanced rapidly, driven by machine learning and Natural Language Processing (NLP) techniques \cite{asim2018ontologylearning}. Recently, Large Language Models (LLMs) have shown promise in supporting ontology learning tasks \cite{babaei2023llms4ol, lo2024end}, especially when combined with human expertise for parts of the ontology development process \cite{garciafernandez2025ontologyengineeringllm}.

Much of the progress in ontology learning so far has concentrated on the lower levels of the ontology learning layer cake, such as term and taxonomy extraction, while relatively little attention has been paid to the automated identification of axioms. Unlike terms and relations, axioms are often implicit and abstract, making them difficult to identify automatically \cite{khadir2021ontologylearning}. Additionally, the flexibility in how axioms can be modelled further complicates the task. Modelling axioms is also a challenging task for humans, as previous studies have pointed out \cite{warren_usability_2014, vigo_overcoming_2014}. Despite these challenges, axioms are important to include since they add to the expressivity and logical consistency of an ontology, enabling inference, validation, and richer reasoning capabilities \cite{corcho2001, armary_ontology_2025}. Being able to automatically identify them, or semi-automatically by giving suggestions, would provide valuable assistance to an ontology developer. Given the broad capabilities of LLMs in handling abstract reasoning and contextual information and their promising results in related tasks, we investigate their ability to identify axioms in ontologies. For ontology engineers, automating such parts of ontology development could lighten the workload and accelerate the process.




In this paper, we investigate automatic axiom identification by LLMs. Specifically, we investigate how well LLMs can infer axioms in ontologies. To this end, we introduce the Ontology Axiom Benchmark \textit{OntoAxiom} that contains nine ontologies and a total of 17.118 triples and 2.771 axioms.  For the experiment with LLMs, we compare two prompting strategies: a Direct approach, where all axioms are predicted at once, and an Axiom-by-Axiom (AbA) approach, where separate prompts are used per axiom type. Our experiments include twelve LLMs and three shot settings. We evaluate the results using precision, recall, and F1 scores. We provide full results and additional data in an open source repository\footnote{\label{footnote:repo}\url{https://gitlab.com/ontologylearning/axiomidentification}}.

This paper is structured as follows: We start by discussing related work on ontology learning, recent developments in applying LLMs to this field, and the role of axioms. In Section \ref{method-ontologies}, we introduce the Ontology Axiom Benchmark by discussing axiom and ontology selection criteria. In Section \ref{method}, we outline our experimental methodology, covering the LLMs, prompt strategies, and evaluation methods. Section \ref{results} presents the results of our experiments. Finally, we provide a discussion of our findings and their implications in Section \ref{discussion}, and we conclude our work in Section \ref{conclusion} and discuss future work.

\section{Related Work} \label{related_work}

In this section, we will describe related work on ontology learning, LLMs, and axiom identification. In Section \ref{relwork:ontologylearning}, we will briefly sketch a history of the field of automatically, or semi-automatically building ontologies and similar models such as knowledge graphs. In Section \ref{relwork:llms4ontologylearning}, we will discuss recent advances in using LLMs for ontology learning tasks. Finally, in Section \ref{relwork:axiomidentification}, we zoom in on related work on automatic axiom identification. 

\subsection{Ontology Learning} \label{relwork:ontologylearning}

Creating ontologies is a time-consuming task which requires the effort of developers and domain experts to properly capture concepts within a domain. The field of ontology learning aims to automate (parts of) this process. Early approaches mostly applied rule-based frameworks \cite{buitelaar2005ontologylearning}, with more recent approaches adopting machine learning techniques \cite{khadir2021ontologylearning}. Giunchiglia et al. \cite{giunchiglia2007lightweight} make a different distinction in complexity. They introduce lightweight ontologies that contain no axioms and complex relations. Heavyweight ontologies, on the other hand, contain more axioms and complex relations \cite{furst2006heavyweight}. The lightweight ontologies were easier to automatically create from text \cite{wong2012ontology}.

With the rise of machine learning techniques and increased computing power, statistical approaches have become more effective in ontology learning. Asim et al. \cite{asim2018ontologylearning} discuss techniques such as co-occurrences, clustering, and word embeddings, comparing linguistic and statistical methods. They also differentiate between term extraction, generally considered the easier task, and relation extraction, considered more complex. Recently, Amalki et al. \cite{amalki_deep_2025} have explored deep-learning techniques such as Recurrent Neural Networks (RNNs), Graph Neural Networks (GNNs) and (knowledge) graph embeddings for ontology learning, including the application of Large Language Models across various domains. They mention an increasing trend in the application of LLMs in the field, as well as a shift towards the use of generalizable, large-scale models. They found that, despite the clear evolution of the field in recent years, challenges remain for smaller domains with less data, particularly in the creation of comparative studies.

\subsection{LLMs for ontology learning} \label{relwork:llms4ontologylearning}

The key technology most current LLMs use is the transformer architecture, which consists of two interacting models, an encoder and a decoder \cite{Vaswani17}. The LLM can be trained end-to-end (such as in Flan-T5 \cite{longpre2023flan}), or using encoder-only architectures (for example BERT \cite{devlin2018bert}) or decoder-only (for example GPT \cite{wan2023gpt-re, gpt4}) models. 

The potential of LLMs has led to significant progress in ontology learning by enabling more accurate and scalable extraction of concepts and relations from unstructured text \cite{amalki_deep_2025}. Examples of popular LLMs in ontology learning are GPT models, Claude, Mistral, BERT, FLAN-T5, Bloom, and RoBERTa \cite{amalki_deep_2025}. Babaei et al. \cite{babaei2023llms4ol} compare the performance of several (Large) Language Models, namely BERT (encoder-only), BLOOM, Llama, GPT-3, GPT-3.5, GPT-4 (all decoder-only), BART and Flan-T5 on the task of Term Typing, Taxonomy Discovery and Non-Taxonomic Relation Extraction. Results show that, for term typing, domain-independent models with large-scale parameters such as GPT and Bloom outperform the other models. For taxonomy discovery, GPT models also outperform the other models, and there is a 10\% performance gap with the open source models. For non-taxonomic relations, Flan was the best model (in 2023). Another result is that the models with the biggest number of parameters have the best results over all tasks, and that, at least for Flan, instruction tuning increases performance on the different tasks.

Beyond step-by-step approaches on different subtasks, researchers have also explored end-to-end ontology learning, where multiple subtasks, such as term and relation extraction, are learned jointly in a single step \cite{lo2024end,bakker2024ontology,lippolis2025ontology}. These efforts have primarily focused on extracting concepts, relations, and taxonomic structures \cite{lo2024end}. A key distinction among these works lies in both the prompting techniques used and the type of input from which ontologies are generated. Inputs range from full-domain texts \cite{bakker2024ontology} to more structured sources like competency questions and short user stories \cite{lippolis2025ontology}. Prompting strategies vary as well, including zero-shot and few-shot approaches (where a small number of labelled examples is included in the prompt), as well as methods that decompose the task, for example, by prompting separately for each competency question \cite{saeedizade_navigating_2024}. Ciatto et al. \cite{ciatto2025large} take a different approach by populating an existing ontology by instantiating classes and properties with LLMs, thereby showing the feasibility of using an LLM as an oracle for ontology population.

Many of these studies are limited to restricted domains, such as music or theatre, where LLMs can use their patterns learned during training to generate the content of the ontologies \cite{lippolis2025ontology}. While these controlled settings provide useful insights, they do not always translate to more specialised or real-world applications \cite{amalki_deep_2025}. For more complex domains, additional domain-specific information can be incorporated, such as news articles \cite{bakker2024ontology} or manually curated domain descriptions \cite{fathallah_llms4life_2024}.

Despite these advancements, significant challenges remain. The lack of standardised benchmarks, the absence of good statistical paradigms to evaluate these benchmarks, the abundance of errors in benchmarks and difficulties in evaluation make it hard to compare different approaches and track progress \cite{lo2024end,amalki_deep_2025,lippolis2025ontology, vaugrante2024}.

\subsection{Axiom Identification} \label{relwork:axiomidentification}
Axioms, being at the top of the ontology learning layer cake in Figure \ref{fig:layer-cake}, are considered a challenging part of ontology learning because they require inferring abstract logical relations rather than surface-level associations. \cite{buitelaar2005ontologylearning,wroblenska2012ol}. There are a few works that specifically focus on automatic axiom identification. Haase et al. \cite{haase2005ontology} take a confidence based approach of ontology learning. For equivalence axioms, they employ a corpus based similarity metric to calculate a confidence score of equality. For disjoint, they implement a heuristic based on lexico-syntactic patterns. This is based on the evidence found for the disjointness in relation to the overall evidence of disjointness throughout all concepts.

B\"uhmann and Lehmann \cite{buhmann2013pattern} mine frequent axiom patterns from over 1300 ontologies and validate their method by manually evaluating them. Their evaluation shows experts deeming 48.2\% of the proposed axioms to be useful extensions to the ontology, indicating the potential usefulness of automated axiom identification for ontology engineering, with room for improvement. 

Ballout et al. \cite{ballout2022predicting} propose a method for axiom learning based on similarity scores derived from ontological distances between concepts. They test these axiom similarity matrices for subClassOf and disjointWith axioms with various regression methods, including random forests and support vector regression.

More recently, LLMs are used to identify axioms from natural language. Mateiu and Groza \cite{mateiu_ontology_2023} finetune a GPT-3 model to convert natural language into structured OWL Functional Syntax. Among other parts of the target ontology such as class subsumption, they include the taxonomical subclass (subClassOf), class disjointness (disjointWith) and class equivalence axioms in their approach. They furthermore prompt GPT-3 to provide domain and range of extracted relations, as well as provide examples of subproperty (subPropertyOf), symmetric and asymmetric relations. However, their evaluation is limited to a brief qualitative assessment of selected examples, making it difficult to fully assess the potential of LLMs for axiom identification.

\subsection{Summary}

Based on the literature, LLMs have led to significant progress in ontology learning \cite{amalki_deep_2025,lo2024end,bakker2024ontology,lippolis2025ontology}. Research has focused on the extraction of concepts, relations and taxonomic structures, but very limited research has been done on axioms \cite{amalki_deep_2025}. To the best of our knowledge, the only research done with axioms and LLMs is by Mateiu and Groza \cite{mateiu_ontology_2023}. This paper explores the potential for the usage of LLMs for axiom extraction. 


\section{Ontology Axiom Benchmark}\label{method-ontologies}

In this paper, we introduce the Ontology Axiom Benchmark \textit{OntoAxiom}. We selected five axioms, which will be discussed in Section \ref{subsec:axiom selection}. Based on these axioms and additional ontology selection criteria, we select nine existing ontologies, which are discussed in Section \ref{subsec:ontology selection}. 

\subsection{Axiom Selection}\label{subsec:axiom selection}

\begin{table}[ht]
\caption{Summary of common ontology axioms, their corresponding description logic and a description of their meaning in natural language.}\label{tab:summary-axioms}
\begin{tabular}{lll}
\toprule
Axiom & D.Logic & Description \\ \hline
subClassOf & $C_1\sqsubseteq C_2$ & All the instances of one class $C_1$ are instances of another class $C_2$\\ 
disjointWith & $C_1\sqsubseteq \neg C_2$ & No instance of one class $C_1$ can be an instance of another class $C_2$\\
\hline
subPropertyOf &$P_1 \sqsubseteq P_2$& All resources related by one property $P_1$ are related by another property $P_2$\\
range &$\exists P .\top \sqsubseteq C$& The values of a property $P$ are instances of class $C$\\
domain & $\top \sqsubseteq \forall P . C$& Any resource that has a given property $P$ is an instance of class $C$ \\
\bottomrule
\end{tabular}
\end{table}

Formal axioms are an important part of ontologies, since they represent universally true knowledge and enforce logical constraints. This allows for reasoning and ensures consistency within an ontology \cite{gruber1993translation}. To evaluate a representative range of axioms, we choose to include both OWL and RDFS axioms, and the choice is made to include both class axioms as well as property axioms.
This leads us to focus on the following five well-known and often used axioms: \triple{rdfs:subClassOf, owl:disjointWith, rdfs:subPropertyOf, rdfs:range}, and \triple{rdfs:domain}. Table \ref{tab:summary-axioms} shows these axioms, their corresponding description logic, and textual definitions. Together, these five axioms form an impactful toolkit to formally model domains with sufficient rules. We introduce the selected five axioms more in-depth in the rest of this section.



\textbf{SubClass and SubProperty.} The subclass (\triple{rdfs:subClassOf}) and subproperty (\triple{rdfs: subPropertyOf}) axioms establish hierarchical relationships between classes and properties, respectively. These axioms allow for inheritance of properties, abstraction and generalisation. An example for the subclass axiom is \triple{:Turtle rdfs:subClassOf :Animal}, which allows to infer \triple{:Turtle} being an \triple{:Animal}. Similarly, for subproperty, an example would be \triple{:ownsTurtle rdfs:subPropertyOf :ownsPet}, which asserts that for any class with \triple{:ownsTurtle} also implies \triple{:ownsPet}. 

\textbf{Disjoint.} Complementing this, the disjoint \triple{OWL:disjointWith} axiom introduces the constraint that asserts two classes cannot share instances, called disjointness. For example, \triple{:Turtle owl:disjointWith :Dog} means that a :Turtle is not a :Dog, something which enforces clarity in distinction and prevents semantic ambiguity. Note that disjointness can also be defined through the \triple{owl:allDisjointClasses} axiom, which identifies classes that are all disjoint from one another. So, \triple{rdf:type owl:AllDisjointClasses; owl:members (:Dog :Cat :turtle).} shows that an instance can only be of \triple{Dog}s, \triple{:Cat}s or \triple{:Turtle}s, and not the others.

\textbf{Domain and Range.} Finally, the domain (\triple{rdfs:domain}) and range (\triple{rdfs:range}) axioms further enrich the ontologies by constraining the subjects and objects that properties can relate to. Continuing the example, \triple{:ownsTurtle rdfs:domain :Person} and \triple{:ownsTurtle rdfs:range :Turtle} enable explicit typing, with any triple using \triple{:ownsTurtle} implies that the subject must be a \triple{:Person} and the object must be a \triple{Turtle}.

\subsection{Ontology Selection}\label{subsec:ontology selection}

\begin{table}[ht]
\caption{Overview of ontologies in the Ontology Axiom Benchmark, with general information, information on class axioms and information on property axioms. The largest amounts are indicated with bold text.}\label{tab:overview-ontologies}

\begin{tabular}{llllllllll}
\toprule
 & ERA & FOAF & GoodRel & gUFO & MO & NS & Pizza & SAREF & Time \\ \hline
 triples & 7021 & 631 & 1834 & 766 & 2141 & 362 & 1848 & 1151 & 1364 \\
classes & 50 & 15 & 37 & 50 & 60 & 35 & \textbf{93} & 31 & 20 \\
properties & \textbf{459} & 60 & 102 & 47 & 165 & 35 & 8 & 71 & 58 \\ \hline
subClassOf & 20 & 10 & 19 & 56 & 59 & 20 & \textbf{80} & 11 & 16 \\
disjointWith & 4 & 8 & \textbf{256} & 29 & 2 & 4 & 5 & 0 & 1 \\
allDisjoint & 0 & 0 & 0 & 6 & 0 & 0 & \textbf{390} & 0 & 0 \\ \hline
subproperty & 25 & 13 & 22 & 14 & \textbf{81} & 0 & 4 & 8 & 13 \\
domain & \textbf{448} & 55 & 96 & 47 & 132 & 1 & 3 & 52 & 55 \\
range & \textbf{456} & 55 & 96 & 43 & 128 & 1 & 4 & 48 & 55\\
\bottomrule
\end{tabular}

\end{table}

To evaluate the capabilities of axiom identification, the benchmark needs to consist of representative existing ontologies. For this, we have collected ontologies that fit the following three criteria: 
\begin{enumerate}
    \item  \textbf{Ontology Size}. The ontologies should be of adequate size to provide proper insight. Ontologies that are too small, while adding to different topics, might yield too little information during evaluation. On the other hand, ontologies can get very large, which is undesirable from a technical limitations standpoint. For this, we have decided to scope the ontology size between 15 and 100 classes, as well as a maximum of 500 properties.
\item  \textbf{Axiom Amount.} The ontologies should have sufficient axioms to test on. In our scope, we are focusing on the \triple{subclass}, \triple{disjoint}, \triple{subproperty}, \triple{domain} and \triple{range} axioms. Ideally, the ontology includes all of these axioms. However, in practice, this is often not the case. Therefore, we have set this criterion to at least 20 and at most 500 instances of one of the five axioms.
\item \textbf{Topic Coverage.} The benchmark should consist of a wide array of topics. Therefore, each selected ontology should introduce semantically different topics and relations when compared to other ontologies in the dataset. Preferably, different types of ontologies are included, such as everyday items, basic descriptions and abstract concepts.
\end{enumerate}

Based on the above criteria, we selected nine different mid-sized ontologies, which comprise the Ontology Axiom Benchmark. The nine ontologies are as follows: 
\begin{itemize}
    \item \textbf{Railways.} The ontology by the European Union Agency for Railways (ERA) \cite{rojas2021LeveragingSemanticTechnologies} describes the European railway infrastructure, with concepts such as \triple{era:Track} and \triple{era:magneticBrakePrevention}. 
    \item \textbf{Persons.} The Friend-of-a-Friend ontology (FOAF) \cite{kalemi2011FOAFAcademicOntologyVocabulary} is a classic example ontology which describes people, their activities, and their relationships. It describes concepts such as \triple{foaf:Group} and \triple{foaf:OnlineChatAccount}. \item \textbf{Products.} The GoodRelations ontology \cite{hepp2008GoodRelationsOntologyDescribing} is used to describe commodity products and services for web sales. It includes concepts such as \triple{gr:WarrantyScope} and \triple{gr:PaymentMethodCreditCard}.
    \item \textbf{Foundational Knowledge.} gUFO \cite{almeida2019gufo} is a lightweight version of the Unified Foundational Ontology (UFO). It holds foundational concepts such as \triple{gufo:NonSortal} and \triple{gufo:Endurant}.  \item \textbf{Music.} The Music Ontology \cite{yves2007music} contains knowledge on the music domain, with concepts such as \triple{mo:SoloMusicArtist} and \triple{mo:Recording}. \item \textbf{Safety.} The Nord Stream Pipeline ontology \cite{bakker2024text} holds knowledge extracted from a news item about the attack on the Nord Stream Pipeline. It contains concepts such as \triple{ns:GasSupply} and \triple{ns:Statement}. \item \textbf{Pizza.} The Pizza ontology is a classic example ontology about pizza, which holds concepts like \triple{:PizzaBase} and \triple{:SauceTopping} \cite{rector2004pizza}. For this paper, we cleaned the Pizza ontology by removing the following contradicting and irrelevant educational example classes: \triple{:CheeseyVegetableTopping, :SpicyPizzaEquivalent, :VegetarianPizzaEquivalent1, :VegetarianPizzaEquivalent2, :UnclosedPizza,} and \triple{:IceCream}). \item \textbf{Smart Appliances.} SAREF \cite{garcia2023etsi} is an ontology for the smart applications domain, with concepts such as \triple{saref:Device} and \triple{saref:FeatureOfInterest}. \item \textbf{Time.} The OWL-Time ontology \cite{pan2006time} describes temporal concepts, such as \triple{time: Friday} and \triple{time:DateTimeDescription}.
\end{itemize}

\vspace{-1em}


An overview of the ontologies in the benchmark is provided in \autoref{tab:overview-ontologies}, where the number of triples, classes, properties, and different axioms is shown for each ontology.  With this, the benchmark spans a variety of topics for a wide coverage: Railways, Products, Persons, Foundational Knowledge, Music, Safety, Pizza, Smart Appliances and Time. In total, the benchmark consists of 17.118 triples, totalling 2.771 axioms. The benchmark is made available through the open-source repository\textsuperscript{\ref{footnote:repo}}.

\vspace{-1em}

\section{Method} \label{method}

\begin{figure}[ht]
    \centering
    \includegraphics[width=.8\textwidth]{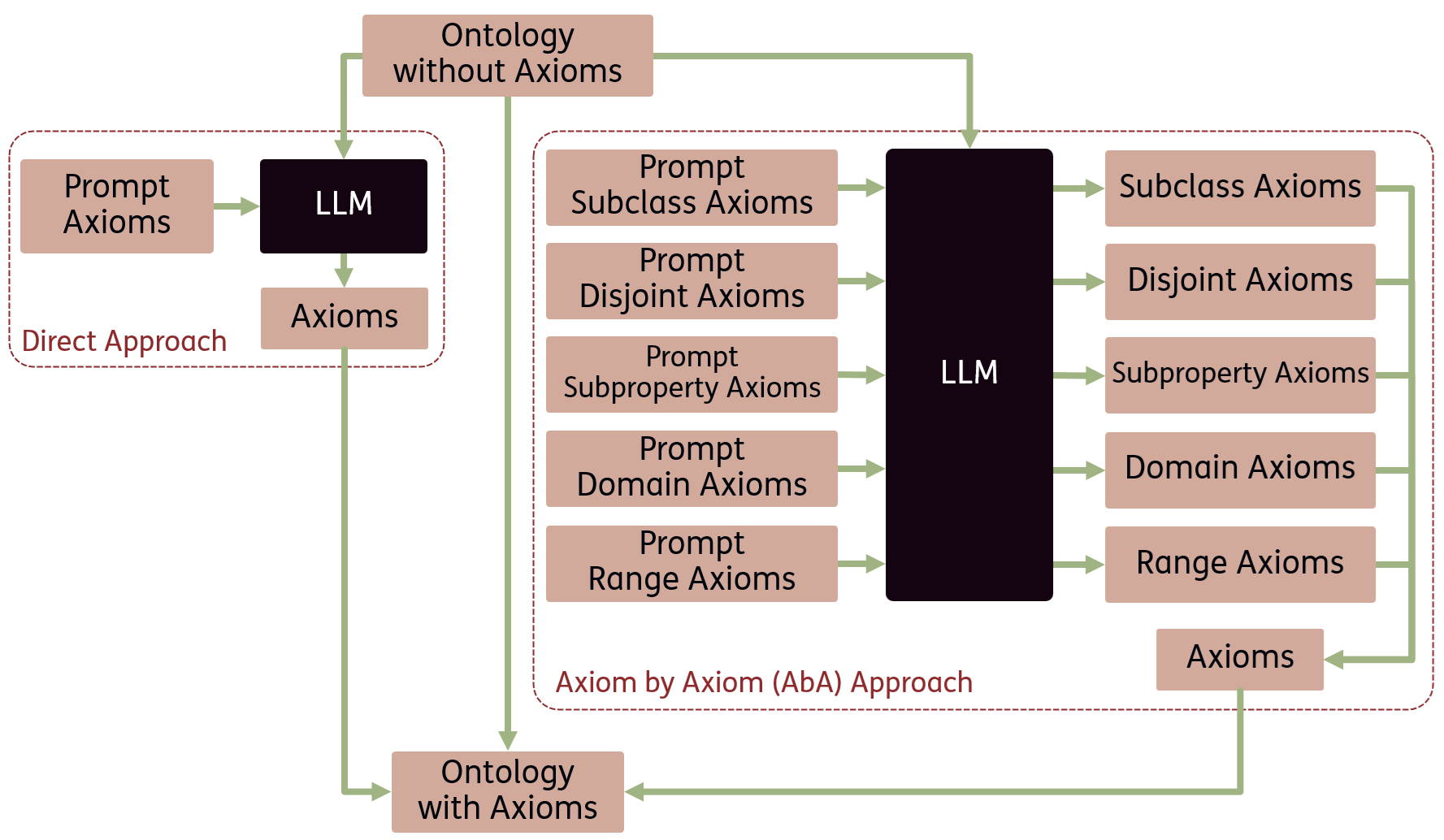}
    \caption{An overview of the two different approaches that we test for axiom identification. Left shows the Direct approach, right the Axiom by Axiom (AbA).}
    \label{fig:Approaches}
    \vspace{-1em}
\end{figure}

This paper explores the ability of LLMs to identify five types of axioms (shown in Table \ref{tab:summary-axioms}). We compare nine ontologies, which were detailed in Section \ref{method-ontologies}. We test two different prompting approaches with three different shot settings, which are described in Section \ref{method-prompts}. We test twelve different LLMs, which are introduced in Section \ref{method-llms}. Finally, we discuss the evaluation strategies in Section \ref{method-eval}. 

\vspace{-2em}

\subsection{Axiom Identification Approaches}\label{method-prompts}

In our experiments, we test two approaches, as shown in Figure \ref{fig:Approaches}. The Figure shows at the top the start of the process with the input: the ontology without axioms. In the middle, the two different approaches are shown, which lead to the end of the process with the output: the ontology with axioms.

In the \textbf{Direct} approach, an LLM is prompted to identify all axioms at once. In the \textbf{Axiom-by-Axiom (AbA)} approach, the axiom identification task is decomposed into smaller steps. The LLM is prompted with separate prompts for each axiom, after which all axioms are aggregated. 

For each of these approaches, we test three shot settings: zero-, one-, and five-shot prompts, an amount frequently chosen for a few-shot setting in previous work \cite{doimo2025representation}. For the zero-shot prompt setting, no examples are included. For the one-shot setting, one labelled example for each axiom is included for the Direct approach, and one example for each separate AbA prompt is included. Finally, for the five-shot approach, five labelled examples are included. The four prompt templates are shown in Table \ref{tab:prompts}: two shot settings for the Direct approach, and two for the AbA approach. For AbA, the prompts for the other axioms are similar to the one shown in the Table, but with different examples. All prompts and included examples can be found in our repository\textsuperscript{\ref{footnote:repo}}.

\begin{table}[ht]
\caption{Prompt templates for both approaches and two shot settings.}
\label{tab:prompts}
\begin{tabularx}{\textwidth}{p{2cm}X}
\toprule
Approach \& Shot setting & Prompt Template \\
\midrule
Direct \newline  Zero-shot &
Identify axioms between these classes and properties.  
The class axioms are subclass and disjointness. The property axioms are subproperty, domain and range.  
Return the output as a JSON object with five keys: \triple{subclass}, \triple{disjoint}, \triple{subproperty}, \triple{domain}, \triple{range}. Each key should map to a list of lists. \newline
 \{CLASSES\}, \{PROPERTIES\}
 \\

Direct \newline  One-shot &
Identify axioms between these classes and properties. These are examples:  
\{subclass: [[Cat, Animal]], disjoint: [[Cat, Dog]], subproperty: [[hasMother, hasParent]], domain: [[Event, happenedAt]], range: [[happenedAt, Place]] \} 
\newline
 \{CLASSES\}, \{PROPERTIES\}
 \\

\midrule
AbA  Zero-shot (Subclass) &
Identify \triple{subclass} relations between these classes: \{CLASSES\}.  
Return an array of arrays, where each inner array contains the subclass first and the superclass second. 
\newline
 \{CLASSES\}, \{PROPERTIES\}
 \\

AbA One-shot (Subclass) &

Identify subclass relations between these classes. This is an example:   
\{[[Cat, Animal], [Couch, Furniture], [Happiness, Emotion]] \} 
\newline
 \{CLASSES\}, \{PROPERTIES\}
 \\
\bottomrule
\end{tabularx}
\vspace{-1em}
\end{table}

\vspace{-1em}

\subsection{LLMs} \label{method-llms}

For our experiments, we selected language models based on whether 1) they are open source or proprietary, 2) what family they are from, 3) what size they are, and 4) their specialisation.  Our goal was to ensure sufficient variety along these dimensions. All models in our selection are instruction-tuned, and for all models we use a temperature setting of 0.2 to ensure a more consistent output. An overview is provided in Table \ref{tab:method_llms}, which also describes the number of parameters. Since the proprietary models from OpenAI are not open source, their parameter counts are not publicly disclosed. We estimated the parameters based on previous work \cite{abacha2024medec}, and categorised the models as large above 30B parameters and small below (blue and orange respectively in Table \ref{tab:method_llms}).

For the open source models, we choose models from the Qwen family, Llama, Mistral and Deepseek. From the Qwen family, we selected Qwen2.5-Coder and Qwen 3. Qwen2.5-Coder is trained on 5.5 trillion tokens of code data \cite{hui2024qwen25}. It is optimised for code generation and reasoning, which might benefit our task. In contrast, Qwen3 is a general-purpose model trained with a Mixture-of-Expert (MoE) architecture \cite{qwen32025technicalreport}. From the Llama family, we choose Llama 3.3, a large general-purpose model from the Llama family \cite{grattafiori2024llama3herdmodels}, as it has been a prominent family of open source LLMs and proven competitive in performance on a variety of tasks \cite{minaee2024large}. Mistral small 3.1 \cite{mistral2025small31}, also a general-purpose model, is included due to its strong performance-to-size ratio; Mistral models have been widely adopted due to efficient inference and high scores on both coding and general reasoning benchmarks at their size class \cite{minaee2024large}. Finally, the last open source model we include is Deepseek R1, a competitive open source reasoning model \cite{deepseekai2025deepseekr1}. 

For the proprietary models, we focus on comparing general purpose models and reasoning models from OpenAI, along with their smaller variants. We include the general purpose models from the OpenAI GPT-4 family GPT-4o and GPT-4o-mini, and GPT-4.1 and GPT-4.1-mini. For the reasoning models we selected o1, o1-mini and o4-mini from the OpenAI-o-family. Previous studies have demonstrated that these models significantly outperform open source models in ontology learning tasks, such as term typing and identifying taxonomical relations \cite{babaei2023llms4ol,doumanas2025llmsOE}. More recently, Lippolis et al. \cite{lippolis2025ontology} found that the o1 model yielded the most promising results, while the Llama-3.1 model exhibited the most structural flaws.

\vspace{-1mm}

\begin{table}[htb]
\centering
\caption{Overview of selected language models (Params colored by size: Blue = Large, Orange = Small.}
\begin{tabular}{p{22mm}llllp{4cm}}
\toprule
Model Name & Family & Params & Specialisation & Open & Exact Model \\
\midrule
Qwen2.5-Coder & Qwen & \cellcolor{TNOOrange}14.8B & Reasoning / coding & Open & Qwen2.5-Coder-14B-Instruct \\
Qwen3         & Qwen & \cellcolor{TNOBlue}30B    & General purpose & Open & Qwen3-30B-A3B \\
Llama 3.3     & Llama 3 & \cellcolor{TNOBlue}70.6B & General purpose & Open & Llama-3.3-70B-Instruct \\
Mistral 3.1 Small & Mistral & \cellcolor{TNOOrange}23.6B & General purpose & Open & mistralai/Mistral-Small-3.1-24B-Instruct-2503 \\
DeepSeek R1   & Deepseek & \cellcolor{TNOBlue}32.8B & Reasoning / coding & Open & deepseek-r1:32b \\
GPT-4o        & GPT4 & \cellcolor{TNOBlue}$\sim$200B & General purpose & Closed & GPT-4o-2024-08-06 \\
GPT-4o-mini   & GPT4 & \cellcolor{TNOOrange}$\sim$8B & General purpose & Closed & GPT-4o-mini-2024-07-18 \\
GPT-4.1       & GPT4 & \cellcolor{TNOBlue}$\sim$1.8T & General purpose & Closed & gpt-4.1-2025-04-14 \\
GPT-4.1-mini  & GPT4 & \cellcolor{TNOOrange}$\sim$8B & General purpose & Closed & gpt-4.1-mini-2025-04-14 \\
o1            & o    & \cellcolor{TNOBlue}$\sim$300B & Reasoning / coding & Closed & o1-2024-12-17 \\
o1-mini       & o    & \cellcolor{TNOBlue}$\sim$100B & Reasoning / coding & Closed & o1-mini-2024-09-12 \\
o4-mini       & o    & \cellcolor{TNOBlue}$\sim$100B & Reasoning / coding & Closed & o4-mini-2025-04-16 \\
\bottomrule
\end{tabular}
\label{tab:method_llms}
\vspace{-1.5em}
\end{table}


\subsection{Evaluation} \label{method-eval}

All results are automatically evaluated with precision, recall, and F1 scores. They are calculated by comparing the predicted axioms with the true axioms in the ontologies. For the disjoint axiom, symmetric predictions are counted as correct, so \triple{A disjoint B} is correct if the ontology states \triple{B disjoint A}. Our experiments cover five axiom types, nine ontologies, two approaches, three shot settings, and twelve LLMs, yielding a total of 3,240 results. In the results section, we present analyses for each of these factors separately, taking the mean over the other factors.


Because LLMs do not consistently output the same results, we add a robustness evaluation, where we run our experiments three times.

\vspace{-0.5em}

\section{Results}\label{results}

In this section, we present the results from our experiments described in the previous section. Section \ref{results:axioms} presents the results on the different axioms. Section \ref{results:ontologies} zooms in on the results of the different ontologies. In Section \ref{results:shotsettings}, we present the results for the direct approach, the AbA (Axiom by Axiom) approach, and the shot settings. Section \ref{results:shotsettings} presents the results over the two approaches and the three different shot settings. Section \ref{results:llms} presents the results for the different LLMs. Finally, we present the results of the robustness evaluation in Section \ref{results:robustness}.

\vspace{-0.5em}

\begin{table}[ht]
\centering
\begin{minipage}{0.45\linewidth}
\centering
\caption{Mean precision, recall, and f1 score per axiom.}
\label{tab:mean-axioms}
\begin{tabular}{lrrr}
\toprule
Axiom & Precision & Recall & F1 \\
\midrule
disjoint & 0.114 & 0.082 & 0.095 \\
domain & 0.034 & 0.043 & 0.038 \\
range & 0.050 & 0.021 & 0.030 \\
subclass & \textbf{0.399} & \textbf{0.326} & \textbf{0.359} \\
subproperty & 0.109 & 0.102 & 0.106 \\
\bottomrule
\end{tabular}
\end{minipage}
\hfill
\begin{minipage}{0.45\linewidth}
\centering
\caption{Mean precision, recall, and f1 score per ontology.}
\label{tab:mean-ontology}
\begin{tabular}{lrrr}
\toprule
Ontology & Precision & Recall & F1 \\
\midrule
ERA & 0.049 & 0.063 & 0.055 \\
FOAF & \textbf{0.257} & \textbf{0.194} & \textbf{0.221} \\
GoodRelations & 0.152 & 0.098 & 0.119 \\
gUFO & 0.208 & 0.144 & 0.170 \\
Music & 0.146 & 0.098 & 0.117 \\
NordStream & 0.084 & 0.122 & 0.100 \\
Pizza & 0.215 & 0.143 & 0.172 \\
SAREF & 0.061 & 0.069 & 0.065 \\
Time & 0.099 & 0.103 & 0.101 \\
\bottomrule
\end{tabular}
\end{minipage}
\end{table}

\vspace{-0.5em}

\subsection{Axioms} \label{results:axioms}

Table \ref{tab:mean-axioms} shows the mean precision, recall, and F1 scores for the different axioms, averaged over the other parameters. The subclass axiom scores highest on all three scores, whereas the range axiom has the lowest F1 score, followed closely by the domain axiom. Precision scores are overall higher than recall, except for the domain. 

\vspace{-1em}


\subsection{Ontologies} \label{results:ontologies}

\vspace{-0.5em}

\begin{figure}[ht]
    \centering
    \includegraphics[width=0.9\linewidth]{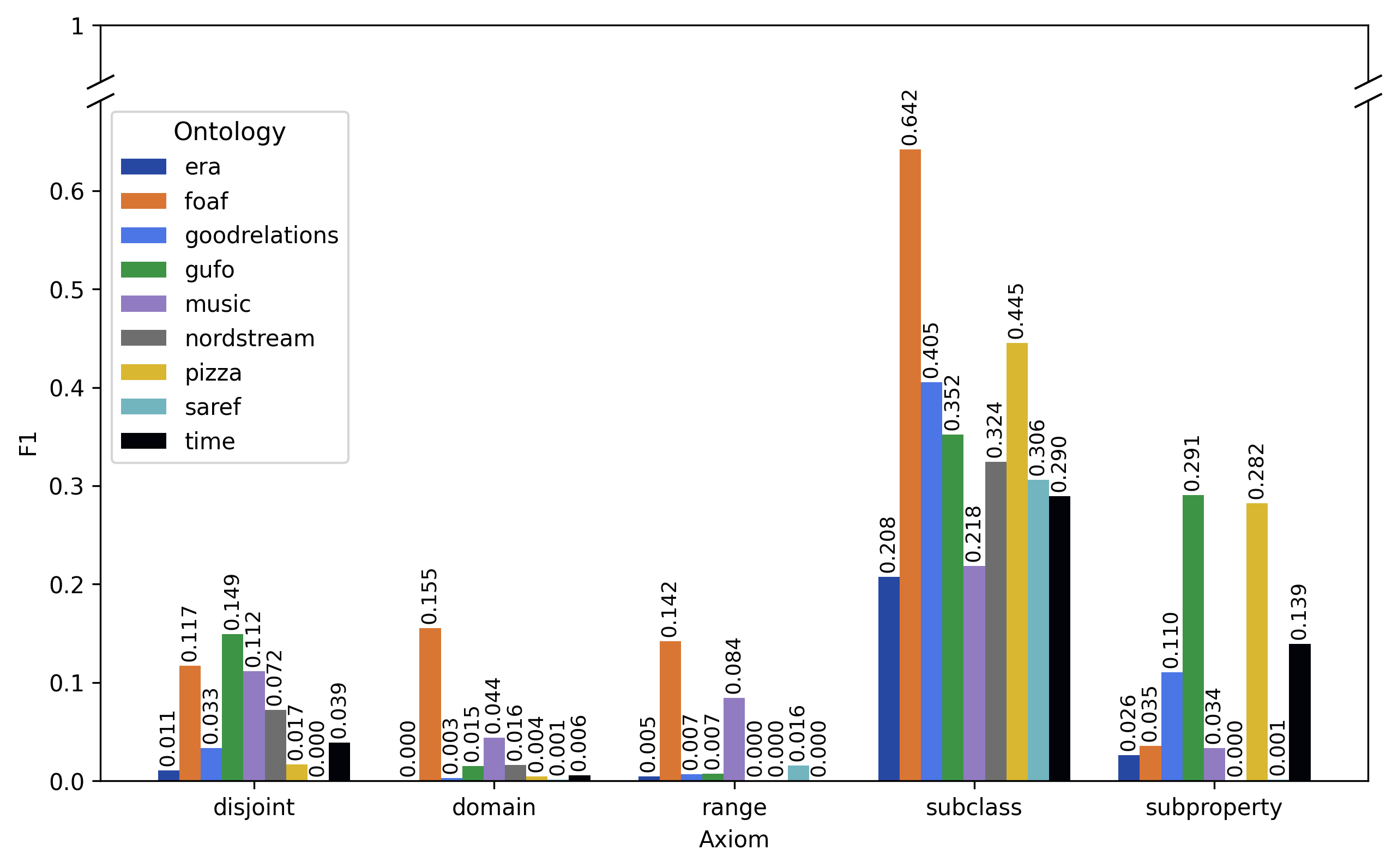}
    \vspace{-1.5em}
    \caption{Average F1 scores per axiom and per ontology.}
    \label{fig:f1_axiom_ontology}
    \vspace{-1em}

\end{figure}


Table \ref{tab:mean-ontology} shows the results of identifying the different axioms per ontology. The ontology with the highest scores is the FOAF ontology, followed by the Pizza and gUFO ontology. The ERA ontology scores lowest. Figure \ref{fig:f1_axiom_ontology} shows the F1 scores per axiom and ontology. Here it can also be observed that the FOAF ontology has the highest scores, although for the subproperty and disjoint axioms, gUFO has higher scores.


\begin{figure}
    \centering
    \includegraphics[width=0.8\textwidth]{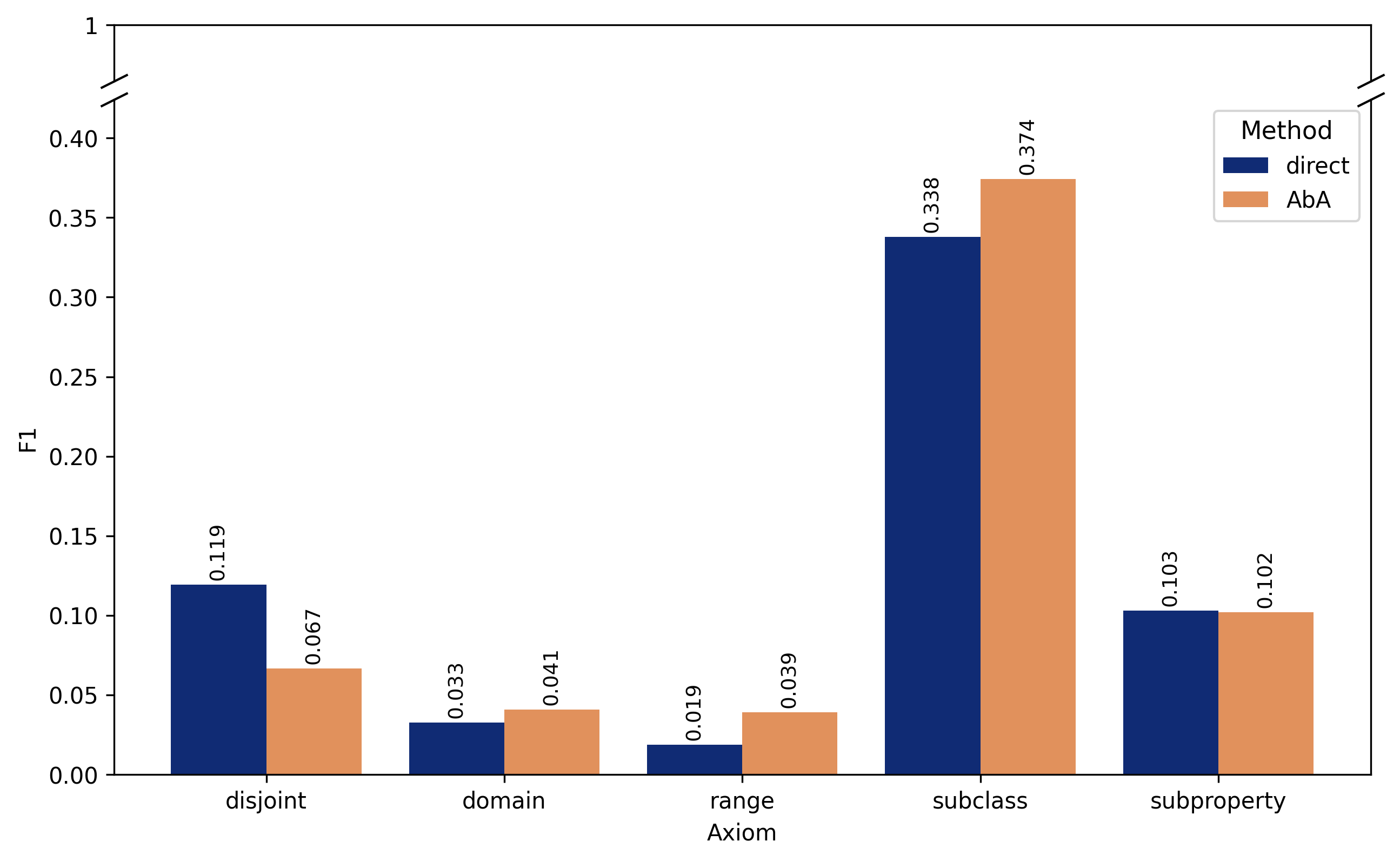}
    \vspace{-1em}
    \caption{Average F1 scores per axiom and per prompting approach.}
    \label{fig:f1_axiom_method}
    \vspace{-0.5em}
\end{figure}

\subsection{Approach \& Shot settings} \label{results:shotsettings}

Table \ref{tab:mean-shot} shows the average F1 scores for the two different prompting approaches and the three shot settings. Overall, the AbA prompting method has the highest F1 score. On the shot level, we can observe the highest scores for the few-shot setting in the direct approach, and for the one-shot setting in the AbA approach. When looking at the different prompting approaches per axiom, shown in Figure \ref{fig:f1_axiom_method}, we can observe that AbA performs better for the subclass, domain, and range axioms. For the disjoint axiom, the direct approach works better, and for subproperty, the performance is almost equal.

\begin{table}[ht]
\centering
\begin{minipage}{0.50\linewidth}
\centering
\caption{Mean precision, recall, and F1 score per approach and shot setting.}
\label{tab:mean-shot}
\begin{tabularx}{\textwidth}{XlXXX}
\toprule
Approach & Shot & Precision & Recall & F1 \\
\midrule
\multirow{3}{*}{direct} 
  & few-shot  & \textbf{0.159} & 0.104 & 0.126 \\
  & one-shot  & 0.158 & 0.099 & 0.122 \\
  & zero-shot & 0.158 & 0.100 & 0.122 \\
\midrule
\multirow{3}{*}{AbA} 
  & few-shot  & 0.125 & 0.127 & 0.126 \\
  & one-shot  & 0.127 & \textbf{0.133} & \textbf{0.130} \\
  & zero-shot & 0.120 & 0.128 & 0.124 \\
\bottomrule
\vspace{1em}
\end{tabularx}
\caption{F1 scores by LLM category (GP = General Purpose, RC = Reasoning / Coding.}
\label{tab:llm-categories}
\begin{tabular}{ll | ll | ll}
\hline
Closed & Open & Large & Small & GP & RC\\
\hline
\textbf{0.124} & 0.077 & \textbf{0.118} & 0.077 & 0.101 & \textbf{0.110} \\
\hline
\end{tabular}

\end{minipage}
\hfill
\begin{minipage}{0.46\linewidth}
\centering
\caption{Mean precision, recall, and F1 score per LLM.}
\label{tab:mean-llm}
\begin{tabularx}{\textwidth}{lrrr}
\toprule
Model & Precision & Recall & F1 \\
\midrule
Qwen2.5-Coder & 0.094 & 0.053 & 0.068 \\
Qwen3         & 0.096 & 0.080 & 0.087 \\
Llama 3.3     & 0.136 & 0.109 & 0.121 \\
Mistral 3.1 Small & 0.106 & 0.108 & 0.107 \\
DeepSeek R1   & 0.130 & 0.060 & 0.083 \\
GPT-4o        & 0.154 & 0.159 & 0.156 \\
GPT-4o-mini   & 0.070 & 0.061 & 0.065 \\
GPT-4.1       & 0.169 & \textbf{0.185} & 0.176 \\
GPT-4.1-mini  & 0.137 & 0.126 & 0.131 \\
o1            & \textbf{0.249} & 0.163 & \textbf{0.197} \\
o1-mini       & 0.146 & 0.099 & 0.118 \\
o4-mini       & 0.206 & 0.176 & 0.190 \\
\bottomrule
\end{tabularx}
\end{minipage}
\end{table}



\subsection{LLMs} \label{results:llms}

Table \ref{tab:mean-llm} shows the mean precision, recall, and F1 score for each LLM. The o1 model scores highest overall (F1 0.164) due to its high precision of 0.249. GPT-4.1 has the highest recall (0.185), and the o4-mini has scores in the same range with an F1 score of 0.197. The models with the lowest scores are GPT-4o-mini, both Qwen models, and the Deepseek model.

Table \ref{tab:llm-categories} shows the performance of the models by their categories as presented in Section \ref{method-llms}: Size, open source or proprietary (Open vs Closed), and specialisation (general purpose or reasoning / coding). Here, it can be observed that the proprietary models by OpenAI perform better overall than the open source models. The large models outperform the small models, and the models specialised in reasoning / coding outperform the general purpose models, although the difference is small in this case (0.009).



\subsection{Robustness} \label{results:robustness}

Table \ref{tab:mean-run} shows the results of the three runs for an indication of how consistent the models perform. Overall, the differences are small, with the precision and recall having very minimal differences in run 1 and 2, and being a bit higher for run 3.



\begin{table}[htb]
\centering
\caption{Mean precision, recall, and f1 score per run.}
\label{tab:mean-run}
\begin{tabular}{lrrr}
\toprule
Run & Precision & Recall & F1 \\
\midrule
run 1 & 0.139 & 0.113 & 0.125 \\
run 2 & 0.139 & 0.114 & 0.125 \\
run 3 & \textbf{0.145} & \textbf{0.118} & \textbf{0.130} \\
\bottomrule
\end{tabular}
\vspace{-1em}
\end{table}




\section{Discussion}\label{discussion}

\paragraph{Axioms} The results for different axioms indicate that subclass axioms are extracted most successfully, whereas domain and range axioms consistently perform worst. This can be partly explained by the fact that, semantically, words for subclass and superclass often lie closer together in meaning; therefore, it might be easier to recognise them. In contrast, difficulties with domain and range axioms might stem from context sensitivity. Unlike subclass axioms, their interpretation can vary depending on the specific context in which the ontology is applied, which likely contributes to the lower extraction performance.


\paragraph{Ontologies}
The domain of the ontology influenced how well axioms could be identified. The FOAF ontology stands out, achieving a score of 0.642 for the subclass axiom (Figure \ref{fig:f1_axiom_ontology}). In contrast, the ERA ontology gives notably lower results. These differences may stem from the domains: the FOAF domain is more general, making axioms easier to identify. Additionally, LLMs are likely trained on data that includes papers on this ontology, since it is an often-used ontology, further improving performance. On the other hand, the ERA ontology is more specialised, containing concepts such as \triple{Siding} and \triple{PhaseInfo}, which may have less publicly available data. 

\paragraph{Approach \& Shot settings}
Prior work has demonstrated the influence of prompting methods on model performance \cite{saeedizade_navigating_2024}. In our experiments, we observe only a modest effect of prompting approach and shot settings. Few-shot and one-shot prompts produced slightly higher F1 scores in some cases, but the differences are limited. This may reflect the intrinsic complexity of axiom extraction: even providing examples does not substantially simplify the task, suggesting that model performance is more constrained by task difficulty than by prompt design.

\paragraph{LLMs}
Overall, the larger LLMs outperformed the smaller ones, with o1 achieving the best results. This is to be expected and in line with previous work; larger LLMs usually outperform the smaller ones, but are also more costly in energy consumption and computing time. Among smaller models, o4-mini performs surprisingly well, achieving a higher F1 score than all models except o1, although precise comparisons are difficult because OpenAI does not disclose model size details.

The proprietary models of OpenAI perform better than the open source models. However, they are more expensive to run, and technical details are not available. Within open source models, Llama 3.3 and Mistral Small perform best, occasionally outperforming some proprietary alternatives, highlighting that careful model selection can mitigate cost without sacrificing performance. Model specialisation had a smaller overall effect, though the two best-performing models (o1 and o4-mini) are reasoning models, suggesting that this task benefits from reasoning capabilities.

Finally, although LLMs can hallucinate and produce inconsistent results, our robustness test shows that, for this task, the differences are small and the general observations still hold. 


Reflecting on this research, the results highlight the difficulty of this task, reflecting the observation that as we move up the ontology learning layer cake, automation becomes increasingly challenging. While LLMs have advanced the extraction of lower-layer knowledge, their effectiveness decreases for higher-layer tasks such as axiom identification. This is not just a limitation of LLMs: developing axioms is also challenging for humans. This suggests that a hybrid approach may be promising. For example, an LLM could generate candidate axioms, which a developer then evaluates and selects according to the specific use case. Alternatively, developers could provide additional documentation or requirements to guide the LLM in generating more relevant axioms. Experimenting with different types of input, such as domain-specific texts or competency questions, could further enhance this collaborative workflow.




\subsection{Limitations}

Even though we extensively compared different prompts, models, and ontologies, there are still aspects that could further improve results. First, we chose to focus on five RDFS/OWL axioms, but extending this work to additional axioms, complex axiom structures, and logical rules could be a valuable direction for future research. Second, we created a benchmark with nine medium-sized ontologies; it would be valuable to extend this benchmark with both more axiom types and ontologies.

In our evaluation, we have chosen a strict scoring method. However, for axioms such as SubClass and SubProperty, more lenient scoring would have been possible. E.g., if `a Pizza is a Food and a Food is a Thing', then the prediction `a Pizza is a Thing' could have been considered correct because of the transitive quality of these axioms. Additionally, we evaluated the results by taking averages over other parameters. When looking at combinations, such as the F1 scores per axiom and per ontology in Figure \ref{fig:f1_axiom_ontology}, more nuances become visible. We highlighted the combinations most relevant to our research goals, such as differences per axioms and per ontology, and axioms and prompting approach. However, more combinations are possible and potentially interesting.

Furthermore, while we spent effort on various settings, other prompting methods could be taken into account, such as Chain of Thought-prompting or varying the few-shot permutations. Finally, evaluating ontologies is inherently challenging, especially for automatically generated ones. Additional evaluation, such as with competency questions or having a human perform this task, could provide deeper insights into the quality and usefulness of the extracted axioms.


\section{Conclusion}\label{conclusion}

In this work, we explored how well LLMs can identify different types of axioms in ontologies. Formal ontologies and axioms are valuable for formalising knowledge within information systems, but their development is resource-intensive. To automate this process, we introduced a benchmark of nine ontologies, 17.118 triples, and 2.771 axioms, and we tested multiple LLMs with different prompting approaches for axiom identification.

Our results indicate that some axiom types, like domain and range, are more challenging to identify than others, and the domain of the ontology significantly impacts performance. Additionally, prompting methods influence results, with the Axiom-by-Axiom prompting method being more successful overall than the Direct prompting approach. Furthermore, the larger LLMs consistently outperform smaller ones. Among those, o1 achieved the highest scores. Notably, for some ontologies and axiom types, F1 scores reached as high as 0.642, demonstrating that LLMs can achieve strong performance on this challenging task.

For future work, it would be interesting to extend this approach to additional axiom types and explore a more interactive workflow between LLMs and developers. This is especially worthwhile for identifying complex axioms, where ontology engineers remain essential. Overall, our findings suggest that LLMs have the potential to support the ontology engineering process. While fully automated axiom identification remains challenging, LLMs can provide valuable candidate axioms to support ontology engineers with the development and refinement of ontologies.

\begin{acks}
    We would like to thank the TrustLLM project for their funding and support for this research. We are also grateful to Cornelis Bouter, who provided us with feedback on the axiom selection and labelled examples for the few-shot experiments.
\end{acks}

\bibliographystyle{SageV}
\bibliography{bibliography}

\end{document}